\documentclass{article}

\usepackage{Styles/iclr2025/iclr2025_conference,times}
\iclrfinalcopy

\usepackage{amsmath,amsfonts,bm}

\def\eqref#1{equation~\ref{#1}}

\def\1{\bm{1}}

\DeclareMathAlphabet{\mathsfit}{\encodingdefault}{\sfdefault}{m}{sl}
\SetMathAlphabet{\mathsfit}{bold}{\encodingdefault}{\sfdefault}{bx}{n}

\usepackage[utf8]{inputenc} 
\usepackage[T1]{fontenc}    
\usepackage{hyperref}       
\usepackage{url}            
\usepackage{booktabs}       
\usepackage{amsfonts}       
\usepackage{nicefrac}       
\usepackage{microtype}      
\usepackage{xcolor}         
\usepackage{graphicx}
\usepackage{caption}
\usepackage{subcaption}
\usepackage{tcolorbox}
\usepackage{algorithm}
\usepackage{algpseudocode}
\usepackage{bigints}

\title{WorldLLM: Improving LLMs' world modeling using curiosity-driven theory-making}

\author{
Guillaume Levy\\
Inria (Flowers AI \& CogSci)\\
Univ. of Bordeaux, France\\
\And
Cédric Colas \\
MIT \\
Inria (Flowers AI \& CogSci) \\
\And Pierre-Yves Oudeyer\\
Inria (Flowers AI \& CogSci)\\
Univ. of Bordeaux, France\\
\And
Thomas Carta\\
Inria (Flowers AI \& CogSci)\\
Univ. of Bordeaux, France\\
\And
Clément Romac\\
Hugging Face \\
Inria (Flowers AI \& CogSci)\\
Univ. of Bordeaux, France\\
}

\newtcolorbox{promptbox}[1]{
  colback=lightgray!30, 
  colframe=black, 
  boxrule=1pt, 
  boxsep=2pt, 
  left=10pt, right=10pt, top=10pt, bottom=10pt, 
  #1
}

\begin{document}

\maketitle

\begin{abstract}
Large Language Models (LLMs) possess general world knowledge but often struggle to generate precise predictions in structured, domain-specific contexts such as simulations. These limitations arise from their inability to ground their broad, unstructured understanding in specific environments. To address this, we present \textbf{WorldLLM}, a framework that enhances LLM-based world modeling by combining Bayesian inference and autonomous active exploration with reinforcement learning. WorldLLM leverages the in-context learning abilities of LLMs to guide an LLM-based world model's predictions using natural language hypotheses given in its prompt. These hypotheses are iteratively refined through a Bayesian inference framework that leverages a second LLM as the proposal distribution given collected evidence. This evidence is collected using a curiosity-driven reinforcement learning policy that explores the environment to find transitions with a low log-likelihood under our LLM-based predictive model using the current hypotheses. By alternating between refining hypotheses and collecting new evidence, our framework autonomously drives continual improvement of the predictions. Our experiments demonstrate the effectiveness of WorldLLM in a textual game environment that requires agents to manipulate and combine objects. The framework not only enhances predictive accuracy, but also generates human-interpretable theories of environment dynamics. 
\end{abstract}

\section{Introduction}
Large Language Models (LLMs) possess broad knowledge about the world, but leveraging this knowledge for precise dynamics modeling remains challenging \citep{vafa_evaluating_2024}. While LLMs can engage in general reasoning, they struggle to make accurate predictions in specific domains with structured observations and dynamics, such as physics simulations or video games. This limitation stems from the gap between their general capabilities and the need for grounded, domain-specific understanding \citep{carta_grounding_2023, cross_hypothetical_2024, bender_climbing_2020}.

The challenge of learning world models has been extensively studied in model-based reinforcement learning (RL), as depicted in \citet{moerland_model-based_2022}. Traditional approaches focus on learning forward dynamics models through direct interactions, typically using neural networks to predict next observations given current states and actions. Recent work has shown promise in using LLMs for world modeling by fine-tuning them on embodied experiences \citep{xiang_language_2023}. However, such an approach is particularly computationally costly given the large amount of interaction data required as well as the cost of gradient-based finetuning of LLMs. Contrary to this, LLMs are known to possess general knowledge that can be adapted using careful prompting \citep{zhang_language-guided_2024}. 

In a parallel line of work, theory-based RL takes inspiration from humans' inductive reasoning and intuitive theory-making to suggest focusing on inferring structured and compositional world models (e.g. programs) \citep{tsividis_human-level_2021}. Current approaches either generate programs to capture environment dynamics directly \citep{tsividis_human-level_2021, tang_worldcoder_2024, li_automated_2024} or use natural language rules as an intermediate representation \citep{piriyakulkij_doing_2024, wang_hypothesis_2023}. Leveraging the world model's structure as well as human-provided prior distributions over possible world models, theory-based methods are usually more sample-efficient than gradient-based ones. More importantly, the resulting world model often shows stronger generalization abilities. However, contrary to gradient-based approaches that usually do not require any expert-given knowledge, theory-based RL methods mostly rely on hand-crafted theory spaces. Finally, one major challenge in learning a world model lies in the data collection. Indeed, collecting transitions that cover the space of possible outcomes in the environment is key for learning an accurate world model. Notably, both classic model-based RL approach and theory-based RL methods agree on the importance of active data collection focusing on evidence with low likelihood under the current model \citep{schmidhuber_curious_1991, pathak_curiosity-driven_2017, burda_exploration_2018, piriyakulkij_doing_2024}.

In this paper, we present \textbf{WorldLLM}, a framework for autonomous improvement of an LLM's world modeling abilities. Our approach combines 1) probabilistic theory induction to produce hypotheses that are given in our LLM's prompt to improve its predictions and 2) curiosity-driven RL to explore the environment and collect transitions poorly predicted by the current hypotheses. Formally, our LLM's world model is the conditional probability $P(s_{t+1}|s_t, a_t, H)$, where $s_t$ represents a state, $a_t$ an action, and $H$ a set of natural language hypothesized theories. This probability is computed by the LLM by giving it $s_t$, $a_t$, and $H$ in its prompt and taking the probability of $s_{t+1}$ to follow this prompt. Our key insight is that natural language theories can help ground an LLM's broad knowledge into precise predictive power by providing domain-specific rules \citep{zhang_language-guided_2024}. Our approach consists of three interacting components: (1) our LLM that computes $P(s_{t+1}|s_t, a_t, H)$ by conditioning its predictions on both a state-action pair and the current hypotheses, (2) a theory generator that updates natural language hypotheses using Bayesian inference, and (3) a curiosity-driven reinforcement learning agent trained to collect evidence against the current hypotheses. Inspired by how humans, from children to scientists \citep{piaget_construction_1954,gopnik_words_1997}, actively update their internal world model by performing experiments, our agent's exploration provides new evidence for hypothesis refinement, creating a virtuous cycle of improvement.


We demonstrate our approach in a video game environment where agents manipulate and combine objects, showing that WorldLLM successfully learns accurate predictive models while generating human-interpretable theories about environment dynamics. This work contributes to a growing body of research on improving LLMs' world modeling capabilities and grounding their knowledge in specific domains. By combining ideas from theory-based RL, Bayesian inference, and active exploration, we provide a framework for learning structured, interpretable world models that leverage both the broad knowledge of LLMs and domain-specific experiences without any costly gradient-based learning.

\begin{figure}[hbt]
    \centering
    \includegraphics[width=0.8\linewidth]{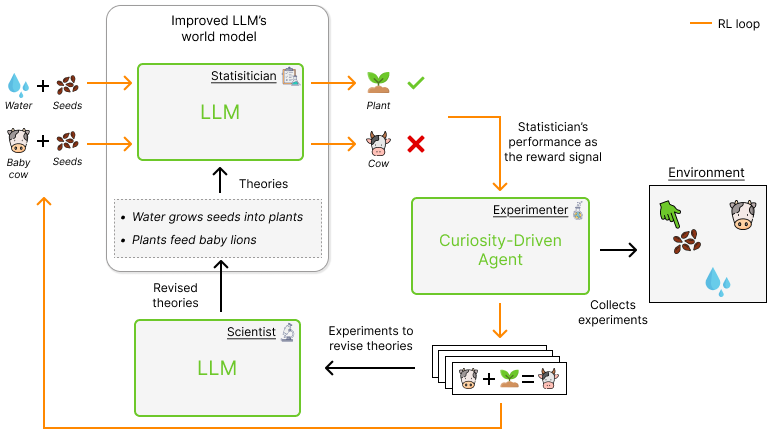}
    \caption{Our framework aims at improving an LLM's ability to predict the outcome of state-action pairs by augmenting it with natural language hypotheses about the world (\textit{Statistician}). WorldLLM alternates between generating hypotheses (\textit{Scientist}) and collecting evidence to update these hypotheses (\textit{Experimenter}). The former uses Bayesian inference with an LLM as the proposal distribution. The latter uses curiosity-driven RL to train an agent to collect experiences that have a low likelihood under the \textit{Statistician} with the current set of hypotheses.}
    \label{fig:4-worldllm-main_schema}
\end{figure}

\section{Improving LLMs' world model with WorldLLM}
\subsection{Problem statement} \label{sec:worldllm-problem_statement}
We consider an environment framed as a Markov Decision Process $\mathcal{M = (S,A,R,T)}$ with $\mathcal{S}$ the state space, $\mathcal{A}$ the action space, $\mathcal{R: S \times A \times S} \mapsto \mathbb{R}$ the reward function, and $\mathcal{T: S \times A \mapsto S}$ the transition function. In particular, as our main objective is to study world modeling abilities, we consider reward-less environments where no particular task must be solved, i.e., the reward function always returns 0. 
We focus on textual environments where, given a vocabulary $\mathcal{V}$ and a maximum sentence length $N$, observations $o \in \mathcal{S} \subset \mathcal{V}^N$ and actions $a \in \mathcal{A} \subset \mathcal{V}^N$ are represented in natural language.

In such a setting, we study how to rapidly learn an LLM-based parametric model $\mathbb{P}^{LLM}$ of the transition function $\mathcal{T}$ (also called forward model) given a budget of experiences $T$ allowed to an agent. This parametric model outputs a probability distribution over all possible next states given a state-action pair. The natural approach from the RL literature to improve this forward model uses gradient-based updates \citep{ha_world_2018, hafner_dream_2020, schrittwieser_mastering_2020}. However, the scale of modern LLMs makes such an approach particularly costly. LLMs' natural in-context learning abilities open up an alternative approach: providing natural language hypotheses in the prompt to improve their predictions. This method is less costly, known as more sample efficient \citep{le_scao_how_2021}, and provides human-interpretable updates. Consequently, the parameter we seek to optimize is a set of hypotheses $\mathcal{H} \subseteq \mathcal{V}^N$ that, when given as input to the forward model (in addition to state-action pairs), maximizes the likelihood of the experienced next states.

We define the global optimization objective as finding the smallest set of hypotheses $H \subseteq \mathcal{H}$ that maximizes the forward model's probability of all possible transitions in the environment:
\begin{equation}
\begin{split}
max & \quad \int_{S\times A\times S} \mathbb{P}^{LLM}(s'|s,a,H) \,ds'\,da\,ds \\
min & \quad |H| \\
\text{subject to} & \quad H \in \mathcal{H} \\
\end{split}
\label{eq:4-worldllm-objective}
\end{equation}

As the full transition space is unknown to our agent, it can only use the collected transitions within the budget of $T$ experiences. Note that it is essential for the agent to efficiently explore the environment within these $T$ experiences, as collecting only similar transitions would lead to hypotheses that predict these transitions well but not the remainder of the transition space. For evaluation, as computing a forward model's performance on the whole transition space is usually intractable, we define a hidden test set $\mathcal{D}_{test}$ containing representative transitions.

\subsection{WorldLLM}
In this section, we present WorldLLM, our methodology for addressing the challenge outlined in the preceding section. Our framework comprises three interconnected components: the Statistician, the Scientist, and the Experimenter. The interactions among these components are illustrated in Figure \ref{fig:4-worldllm-main_schema}. 
The Statistician represents our LLM-based forward model (i.e., $\mathbb{P}^{LLM}$) and is used to evaluate the  current hypotheses. The set of hypotheses $H$ the Statistician uses is produced by the Scientist, another LLM. These hypotheses are derived from trajectories collected from interactions with the environment using the Experimenter. WorldLLM alternates from during $T$ iterations between evidence collection from the Experimenter and hypotheses refinement from the Scientist. In the subsequent sections, we elaborate on how these modules contribute to our objective's optimization.

\paragraph{Statistician}
We first detail the Statistician module, whose role is to evaluate a set of hypotheses $H_t$ on collected transitions. We start by renaming, for the sake of clarity, $\mathbb{P}^{LLM}$ by $\mathbb{P}^{St}$. As described above, the Statistician is a forward model computing the likelihood $\mathbb{P}^{St}(s'|s,a,H_t)$ of the next state $s'$ given a state-action pair $(s,a)$ and a set of hypotheses $H_t$. This likelihood is obtained by the LLM by giving it $s$, $a$, and $H_t$ in the prompt and taking the probability of $s'$ to follow this prompt.

\paragraph{Scientist}
To improve the Statistician's prediction as a forward model, the Scientist's role is to search through the space of possible sets of hypotheses. To efficiently explore this space, we take inspiration from \citep{wang_hypothesis_2023,piriyakulkij_doing_2024} and use a Bayesian inference framework. In particular, we rely on another LLM $\mathbb{P}^{Sc}$ as the proposal distribution which generates a candidate set of hypotheses following the distribution $\mathbb{P}^{Sc}(\hat{H}_t^i|\mathcal{D}_t,\hat{H}_t)$ given a data set of collected transitions $\mathcal{D}_t$ and the current best hypotheses candidate $\hat{H_t}$. We assume a uniform prior (i.e., no particular prior) over the space of hypotheses. For this paper, we chose the Metropolis algorithm, a variant of the Metropolis-Hastings algorithm, which iteratively generates candidate hypotheses for $n_steps$. The full algorithm is displayed in Algorithm \ref{alg:4-worldllm-main} and more details are provided in Appendix \ref{app:worldllm}. 

\begin{algorithm}
\caption{WorldLLM}
\begin{algorithmic}[1]
\State \textbf{Input:} $T$ total number of iterations, $n_{steps}$ number of Metropolis steps at each iteration, $\mathbb{P}^{St}$ the Statistician's distribution, $\mathbb{P}^{Sc}$ the Scientist's proposal distribution, $\pi$ the Experimenter
\State \textbf{Initialize} $H_t \leftarrow null$ 
\For{$t=0$ to $T$}
    \State Collect trajectories $\mathcal{D}_t \leftarrow$env$(\pi)$
    \State Evaluate current hypotheses $current\_score = \sum\limits_{(s,a,s')\sim \mathcal{D}_t}\log\mathbb{P}^{St}(s'|s,a, H_t)$ 
    \State Set $\hat{H}_t = H_t$
    \For{$i=0$ to $n_{steps}$}\Comment{Metropolis step}  
    \State Generate hypotheses $\hat{H}_t^i \sim\mathbb{P}^{Sc}(\cdot|\mathcal{D}_t,\hat{H}_t)$
    \State Evaluate candidate hypotheses $score = \sum\limits_{(s,a,s')\sim \mathcal{D}_t} \log\mathbb{P}^{St}(s'|s,a, \hat{H}_t^i)$
    \State Generate a random number $u \sim \mathcal{U}(0,1)$ 
    \If{$\log(u) < score - current\_score$} \Comment{Accept or reject}
    \State Update hypothesis $\hat{H}_t = \hat{H}_t^i$
    \State Update score $current\_score = score$
    \EndIf
    \EndFor
    \State Set $H_t = \hat{H}_t$
    \State Update Experimenter $\pi$ 
\EndFor
\end{algorithmic}
\label{alg:4-worldllm-main}
\end{algorithm}

\paragraph{Experimenter} \label{sec:worldllm-experimenter-details}
Given a set of hypotheses produced by the Scientist, the Experimenter must collect transitions that will help assess and refine the hypotheses (i.e., transitions that have a low likelihood under the Statistician). We propose to explore two types of approaches. First, we design oracle agents, allowing us to disentangle the efficiency of data collection and hypotheses refinement in WorldLLM.

For the oracles, we propose the following hard-coded policies:
\begin{itemize}
    \item A policy that collects the same distribution of transitions as the test set's distribution (O-Ideal)
    \item An optimal policy that progressively moves from collecting simple transitions to collecting the hardest ones (O-Curriculum)
    \item An optimal policy that produces the sequence of actions necessary to collect the hardest transitions (O-Hardest)
\end{itemize}
While O-Hardest will collect complex transitions, it may not cover the full transition space. In comparison, O-Ideal matches the test set distribution. However, the iterative process of WorldLLM may require the Experimenter to focus for several iterations on particular subspaces of the transition space for the Scientist to produce accurate hypotheses. While O-Hardest and O-Ideal lack such a developmental trajectory, O-Curriculum focuses on specific transitions for an arbitrary number of collections. We also study the performance of a policy acting randomly (O-Random), which is unlikely to collect complex transitions.

Then, we introduce RL-based Experimenters. We propose to study three curiosity-based intrinsic rewards derived from $\mathbb{P}^{St}$ \citep{oudeyer_what_2007}. First, we introduce \textit{RL-LogP}, where the Experimenter uses the prediction error \citep{schmidhuber_curious_1991, pathak_curiosity-driven_2017, burda_large-scale_2018} as reward by receiving $r=-log\mathbb{P}^{St}(s'|s,a,H_t)$ for each collected transition $(s,a,s')$. While usually efficient, this method is also known to suffer in stochastic environments where it fails to separate epistemic from aleatoric uncertainty \citep{burda_exploration_2018}. Second, we propose \textit{RL-ALP}, which leverages information gain over the prediction error (a form of Learning Progress) \citep{oudeyer_what_2007, lopes_exploration_2012, schmidhuber_curious_1991} and rewards transitions on which the forward model improves. In particular, we use the absolute value of Learning Progress, which tracks both progress and forgetting in the forward model \citep{oudeyer_what_2007, baranes_r-iac_2009}:
\begin{equation}
    r_{alp} = |\log\mathbb{P}^{St}(s'|s,a,H_{t-1}) -\log\mathbb{P}^{St}(s'|s,a,H_t)|
    \label{eq:4-worldllm-RL-ALP}
\end{equation} 

Finally, directly tracking Learning Progress at the transition level often leads to noisy estimations. We also propose \textit{RL-ALPEXP}, a variant of RL-ALP which partitions the transition space into subspaces and computes ALP over each subspace, stabilizing the estimations. While existing works studied how to automatically infer such partitions \citep{baranes_active_2013}, we here rely on hand-designed ones, which correspond to the partitions used by oracles. Further details are provided in Appendix \ref{app:experimental_setup}.

\section{Experiments}
\subsection{Experimental setup}
\paragraph{Environment}
\begin{figure}[hbt]
    \centering
    \includegraphics[width=0.8\linewidth]{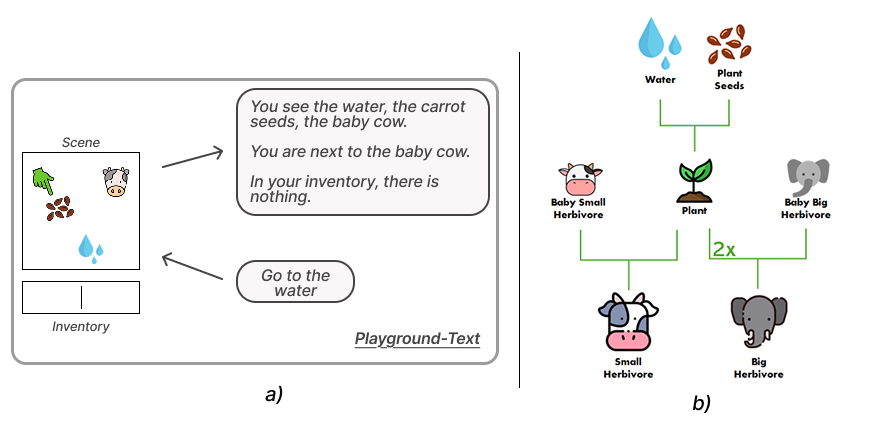}
    \caption{\textit{Our experimental setup}. We use the Playground-Text environment that features objects the agent can interact with, along with textual observations and actions (a). We focus on 4 object types from Playground-Text: \textit{Water}, \textit{Plants}, \textit{Small Herbivores}, and \textit{Big Herbivores}. We show in (b) the technology tree for combining objects.}
    \label{fig:4-worldllm-playground_env}
\end{figure}
To evaluate WorldLLM's efficiency, we use the Playground-Text environment from \citep{gaven2024sacglamimprovingonlinerl}, a textual environment which produces textual observations and receives textual commands. This environment generates a room that contains multiple objects the agent can interact with by grasping or releasing them. The agent can combine two objects, if an interaction is possible, by releasing one onto the other, as illustrated in Figure \ref{fig:4-worldllm-playground_env} b. For instance, releasing \textit{water} on a \textit{carrot seed} transforms them into \textit{grown carrot}. In our experiments, we only consider 4 types of objects: \textit{Water}, \textit{Plants} (seed or grown) that require \textit{water} to grow, \textit{Small Herbivores} (baby or grown) that require a grown \textit{plant} to grow, and \textit{Big Herbivores} (baby or grown) that require 2 grown \textit{plants} to grow. We evaluate WorldLLM's ability to discover, predict, and explain how to grow \textit{Plants}, \textit{Small Herbivores}, and \textit{Big Herbivores}. We provide more details on the environment in Appendix \ref{app:experimental_setup}. For analysis, we gather the possible transitions into 6 different types: \textit{Standing} when moving on an object, \textit{Holding 1} when grasping an object, \textit{Holding 2} when grasping two objects at the same time, \textit{Grow Plant} when releasing water on a seed, \textit{Grow S. Herbivore} when releasing a grown plant on a baby small herbivore and \textit{Grow B. Herbivore} when releasing two grown plants on a baby big herbivore. These types are the ones used by RL-ALPEXP.

\paragraph{Optimization} We use Phi-3-mini-4k-Instruct\footnote{https://huggingface.co/microsoft/Phi-3-mini-4k-instruct} for both the Statistician and the Scientist. For our Experimenter, we either use one of our hard-coded oracles or an RL agent. Regarding oracles, O-Ideal collects transitions that match the distribution of the above test set. O-Hardest performs the optimal policy to grow the \textit{Small Herbivore} and the \textit{Big Herbivore} accessible. O-Curriculum plays the optimal policy to grow a \textit{Plant} up to 133 iterations, then plays the optimal policy to grow the accessible \textit{Small Herbivore} up to 266 iterations, and finally grows the \textit{Small Herbivore} and the \textit{Big Herbivore} until the end. See Appendix~\ref{app:heuristic-agents} for more details on the oracles. Concerning the RL-based Experimenters, while it would appear natural to use the LLM already leveraged in the Statistician and Scientist as the policy (e.g., with GLAM), we propose a simpler setup to focus on world modeling abilities. We use a Multi-Layer Perceptron with $2$ hidden layers of $64$ units as the policy, which uses a symbolic representation of Playground-Text's state as observation. In all experiments below, unless specified otherwise, we use $8$ different random seeds. For each seed, we perform $400$ iterations of the framework (i.e., $T$) where $150$ transitions are collected per iteration and $n_{steps}=5$ steps of the Metropolis algorithm are performed by the Scientist at each iteration. When the Experimenter is an RL agent, we collect $3600$ transitions per iteration to train it and use PPO to update the policy, while only keeping the last $150$ transitions for the Scientist to match the oracles' hyperparameters. Our test set $\mathcal{D}_{test}$ is composed of 120 \textit{Standing}, 20 \textit{Holding 1}, 7 \textit{Holding 2}, 12 \textit{Grow Plant}, 6 \textit{Grow S. Herbivore} and 3 \textit{Grow B. Herbivore}.

\begin{figure}[hbt]
    \centering
    \includegraphics[width=0.9\linewidth]{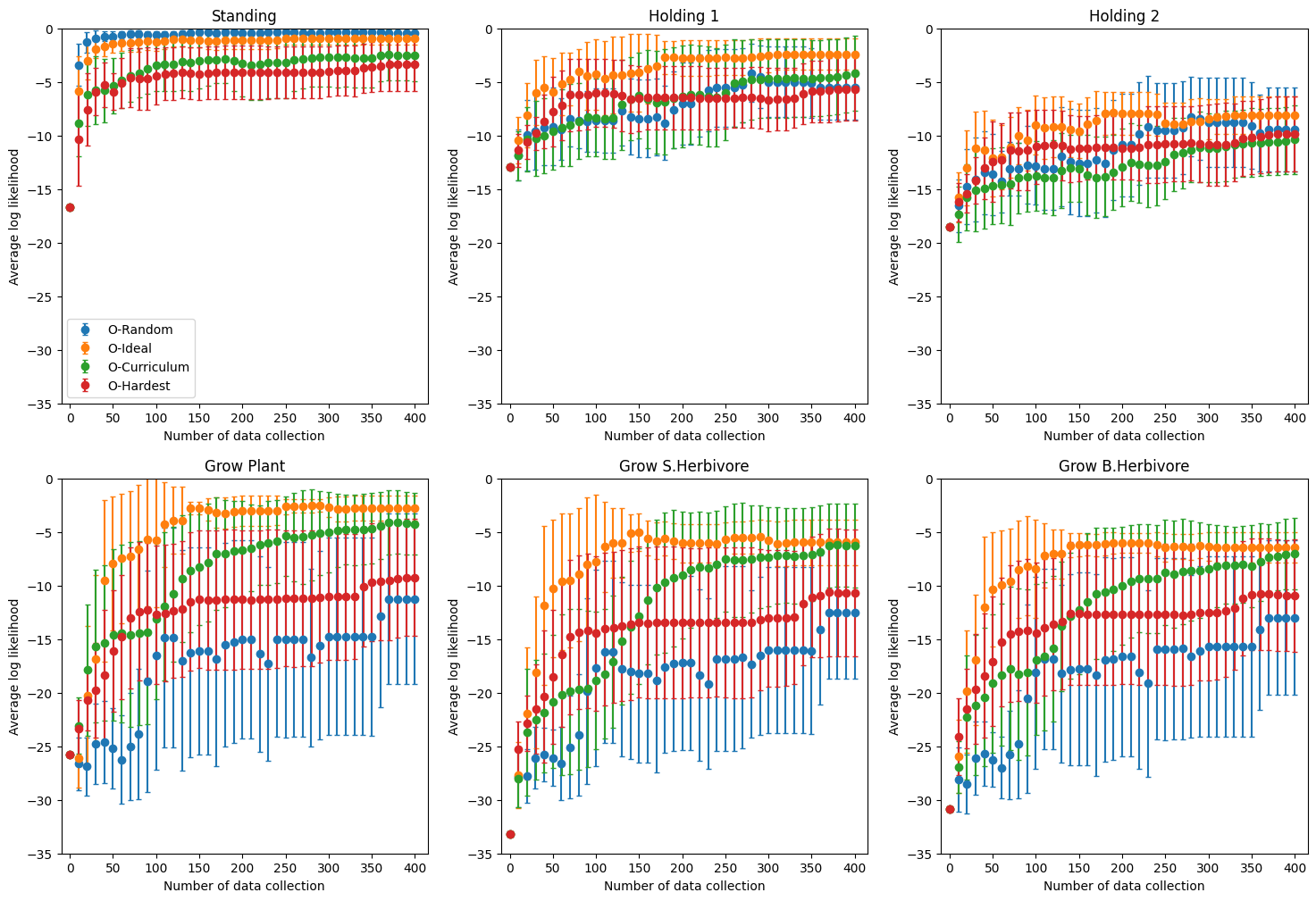}
    \caption{Evolution of log-likelihoods computed by the Statistician (using the Scientist's last accepted hypotheses) on the test set throughout WorldLLM's iterations when using different oracles as Experimenters. We average per transition type over $8$ random seeds with the standard deviation represented by the error bars. We show at iteration 0 the log-likelihood from the Statistician without any hypotheses given.}
    \label{fig:4-worldllm-LogLikelihoodOracle}
\end{figure}

\subsection{WorldLLM with an oracle Experimenter}
We initiate the experiments by examining the performance of WorldLLM when using our oracles as Experimenters. In Figure \ref{fig:4-worldllm-LogLikelihoodOracle}, we show the evolution of log-likelihoods outputted by the Statistician using the last accepted set of hypotheses $H_t$ throughout WorldLLMs' iterations when using different oracles as Experimenters. These log-likelihoods are computed on the test set and averaged over transition types (and seeds).

One can observe that, regardless of the oracle used or transition type, giving the Statistician hypotheses always improves its performance compared to when no hypotheses are given (i.e., the initial point of each curve). When comparing the oracles, one can observe that the data collection strategy of O-Ideal strongly influences WorldLLM's abilities at producing efficient hypotheses. Indeed, using the same distribution as the one we perform evaluation on, O-Ideal achieves the best overall performance. On the opposite, O-Random obtains the poorest performance on complex transitions (i.e., growing an object) while being the best method on \textit{Standing} transitions. This can easily be explained by the fact that moving actions are predominant in the action space. O-Hardest did not lead to strong performance, highlighting the need of diversity in collected transitions for the Scientist. Finally, O-Curriculum obtains results that are on par with those of O-Ideal. Results from Appendix~\ref{app:RetentionRate} indicate that O-Curriculum particularly helps the Scientist generate improved hypotheses.

Overall, these initial experiments demonstrate that WorldLLM is able to produce hypotheses that improve the predictive capabilities of the Statistician's LLM. Moreover, the different performance obtained by our oracles highlighted the importance of 1) diversity in the collected transitions and 2) a developmental collection strategy allowing the Scientist to progressively update the hypotheses. 

\subsection{WorldLLM with a curiosity-driven RL agent as Experimenter} \label{sec:worldllm-rl}
We now move to curiosity-driven RL agents as Experimenters. Compared to the hard coded policies, we rely instead on one of our three intrinsic reward signals (RL-ALPEXP, RL-LogP, and RL-ALP), to collect transitions that drive the Scientist towards generating efficient hypotheses.

As in the previous section, we analyze the evolution of the Statistician's log-likelihoods over the test set. In order to ease the comparison of our three RL-based Experimenters to oracles, we introduce a coverage metric as the area under the evolution of log-likelihoods' curve of each method. We normalize this area using the area between the Statistician's log-likelihood without any hypotheses and the best performance achievable (i.e., 0). The complete formula is
\begin{equation}
    \text{area} = 1-\bigintssss_0^{T}\frac{\log\mathbb{P}^{S_t}(\mathcal{D}_{test}|H_t)}{\log\mathbb{P}^{S_t}(\mathcal{D}_{test}|\{\emptyset\}) T}dt.
    \label{eq:4-worldllm-auc}
\end{equation}

The best performing method possible is expected to be very close to $1$ while an inept method's result will be near $0$. A method whose result is below $0$ means that the hypotheses degrade the initial performance of the Statistician. 

\begin{table}[hbt]
\centering
{\tiny
\begin{tabular}{lcccccc}
\toprule
\textbf{Methods} & \textbf{Standing} & \textbf{Holding 1} & \textbf{Holding 2} & \textbf{Grow Plant} & \textbf{Grow S.H.} & \textbf{Grow B.H.} \\
\midrule
O-Random & 0.97 $\pm$ 0.01 & 0.45 $\pm$ 0.16 & 0.40 $\pm$ 0.15 & 0.34 $\pm$ 0.24 & 0.44 $\pm$ 0.16 & 0.40 $\pm$ 0.17\\
O-Ideal & 0.92 $\pm$ 0.03 & 0.72 $\pm$ 0.12 & 0.50 $\pm$ 0.06 & \textbf{0.81 $\pm$ 0.06} & \textbf{0.77 $\pm$ 0.06} & \textbf{0.74 $\pm$ 0.04} \\
O-Curriculum & 0.79 $\pm$ 0.15 & 0.48 $\pm$ 0.25 & 0.31 $\pm$ 0.15 & 0.67 $\pm$ 0.15 & 0.63 $\pm$ 0.13 & 0.60 $\pm$ 0.14\\
O-Hardest & 0.74 $\pm$ 0.13 & 0.48 $\pm$ 0.20 & 0.39 $\pm$ 0.16 & 0.52 $\pm$ 0.21 & 0.57 $\pm$ 0.18 & 0.56 $\pm$ 0.18\\
\midrule
RL-ALPEXP & 0.95 $\pm$ 0.02 & 0.44 $\pm$ 0.16 & 0.40 $\pm$ 0.14 & 0.42 $\pm$ 0.25 & 0.52 $\pm$ 0.19 & 0.49 $\pm$ 0.19\\
\midrule
RL-LogP & 0.93 $\pm$ 0.01 & \textbf{0.52 $\pm$ 0.07} & \textbf{0.53 $\pm$ 0.07} & 0.62 $\pm$ 0.09 & 0.61 $\pm$ 0.07 & 0.60 $\pm$ 0.07\\
RL-ALP & \textbf{0.98 $\pm$ 0.01} & 0.18 $\pm$ 0.08 & 0.17 $\pm$ 0.06 & -0.07 $\pm$ 0.05 & 0.16 $\pm$ 0.04 & 0.08 $\pm$ 0.07\\
\bottomrule
\end{tabular}}
\caption{This table presents the normalized area under the training curve for each algorithm and transition type. The normalization is performed by dividing the computed area by the area formed by the initial log-likelihoods, i.e., obtained without using hypotheses.}
\label{tab:4-worldllm-loglikelihood_auc}
\end{table}

From the results shown in Table~\ref{tab:4-worldllm-loglikelihood_auc}, one can observe that RL-ALPEXP is able to leverage its environment knowledge to reach performance significantly better than O-Random but remaining worse than O-Ideal. When given no expert knowledge, RL-ALP achieves poor performance while RL-LogP obtains results close to or even better than oracle baselines.

\begin{figure}[hbt]
    \centering
    \includegraphics[width=0.32\linewidth]{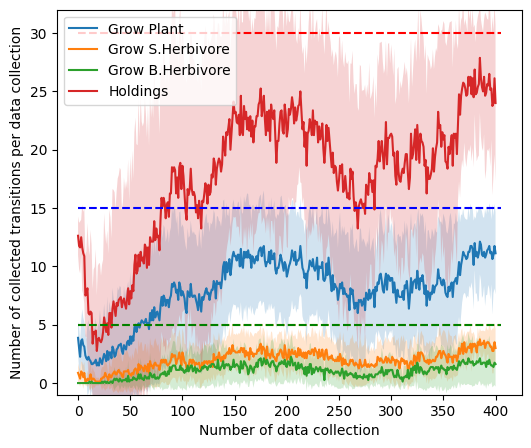}
    \includegraphics[width=0.32\linewidth]{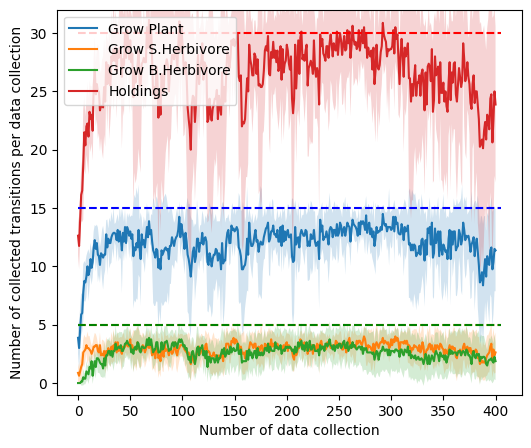}    
    \includegraphics[width=0.32\linewidth]{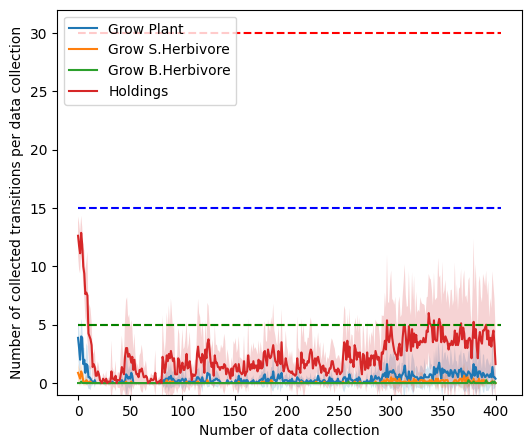}   
    \caption{Proportion of collected transitions for RL-ALPEXP (left), RL-LogP (middle), and RL-ALP (right). The colored dashed lines correspond to the amount of transitions collected by O-Hardest.}
    \label{fig:4-worldllm-RLProportionTransition}
\end{figure}

To better understand the effect of the RL-based Experimenters on the Statistician's predictions, we studied the evolving distribution of collected transitions. We grouped \textit{Holding 1} and \textit{Holding 2} for simplicity and ignored \textit{Standing} transitions, which are predominant by nature. We show these results in Figure~\ref{fig:4-worldllm-RLProportionTransition}.

They show that RL-LogP manages to explore the environment and cover the whole transition space. The resulting collected distribution is near O-Hardest's one (which is shown in dashed lines). This performance is notably explained by the syntax used by Playground-Text for observations and actions that make complex transitions naturally harder to predict for an LLM (e.g., moving to a \textit{Big Herbivore} involves only this entity whereas growing a \textit{Big Herbivore} implies an action where two \textit{Plants} are released on a baby \textit{Big Herbivore}). This natural ordering appears in the rewards obtained by the RL policy and favors complex transitions. RL-ALPEXP, which does not leverage such a reward ordering, progressively converges to collecting more complex transitions. When looking at RL-ALP, most of the collected transitions are \textit{Standing} ones. As we have seen in the previous section, the diversity of the transitions given to the Scientist is key in WorldLLM. Consequently, the hypotheses produced with RL-ALP focus solely on explaining \textit{Standing} transitions. A more detailed analysis of why RL-ALP failed is available in Appendix~\ref{app:rl-alp-failed}.

This section's results show that WorldLLM successfully improves the LLM's forward modeling abilities when using curiosity-driven RL-based Experimenters. Moreover, we show that rewarding Experimenters for collecting transitions that help refine the hypotheses not only leads to efficient hypotheses but also leads to policies that solve our environment's full technology tree. 

\subsection{Evaluating the produced world model} \label{sec:worldllm-eval_world_model}
We now focus on evaluating the usefulness of the produced world model (i.e., the Statistician equipped with the last accepted hypotheses). While previous sections indicate that WorldLLM succeeds in increasing the likelihood of the observed next states, these results do not guarantee that the observed next states receive the highest likelihood compared to other possible next states. We thus examine our LLM-based world model's capacity to correctly predict (i.e., generate) the next state on our test. In particular, for each state-action pair in $\mathcal{D}_{test}$, we used our Statistician's LLM and constrained decoding to only generate the valid next states, given the current state, in Playground-Text. 
We then report in Table~ \ref{tab:4-worldllm-constrained-decoding} the percentage of next states that belonged to the top-3 generated sequences.

\begin{table}[ht!]
\centering
{\tiny
\begin{tabular}{lcccccc}
\toprule
\textbf{Methods} & \textbf{Standing} & \textbf{Holding 1} & \textbf{Holding 2} & \textbf{Grow Plant} & \textbf{Grow S.H.} & \textbf{Grow B.H.} \\
\midrule
No Hypo. & 0.82 $\pm$ 0.00 & \textbf{1.00 $\pm$ 0.00} & 0.00 $\pm$ 0.00 & 0.00 $\pm$ 0.00 & 0.00 $\pm$ 0.00 & 0.00 $\pm$ 0.00 \\
\midrule
O-Random & \textbf{1.00 $\pm$ 0.00} & 0.48 $\pm$ 0.26 & 0.07 $\pm$ 0.19 & 0.18 $\pm$ 0.31 & 0.00 $\pm$ 0.00 & 0.04 $\pm$ 0.11 \\
O-Ideal & \textbf{1.00 $\pm$ 0.00} & 0.94 $\pm$ 0.15 & 0.00 $\pm$ 0.00 & \textbf{0.94 $\pm$ 0.17} & 0.25 $\pm$ 0.26 & 0.46 $\pm$ 0.37 \\
O-Curriculum & \textbf{1.00 $\pm$ 0.00} & 0.88 $\pm$ 0.31 & 0.00 $\pm$ 0.00 & 0.85 $\pm$ 0.28 & \textbf{0.60 $\pm$ 0.41} & \textbf{0.54 $\pm$ 0.44} \\
O-Hardest & \textbf{1.00 $\pm$ 0.00} & 0.86 $\pm$ 0.21 & 0.11 $\pm$ 0.28 & 0.29 $\pm$ 0.34 & 0.27 $\pm$ 0.37 & 0.21 $\pm$ 0.37 \\
\midrule
RL-ALPEXP & \textbf{1.00 $\pm$ 0.00} & 0.71 $\pm$ 0.29 & \textbf{0.34 $\pm$ 0.36} & 0.40 $\pm$ 0.43 & 0.02 $\pm$ 0.06 & 0.08 $\pm$ 0.14 \\
\midrule
RL-LogP & \textbf{1.00 $\pm$ 0.00} & 0.67 $\pm$ 0.30 & \textbf{0.34 $\pm$ 0.25} & 0.74 $\pm$ 0.32 & 0.02 $\pm$ 0.06 & 0.29 $\pm$ 0.42 \\
RL-ALP & \textbf{1.00 $\pm$ 0.00} & 0.20 $\pm$ 0.26 & 0.00 $\pm$ 0.00 & 0.03 $\pm$ 0.08 & 0.00 $\pm$ 0.00 & 0.00 $\pm$ 0.00 \\
\bottomrule
\end{tabular}}
\caption{Top 3 performance of Constrained Decoding for each algorithm by transition type. We also report the performance without any hypotheses given (No Hypo.).}
\label{tab:4-worldllm-constrained-decoding}
\end{table}

The results show that no method performs best on all transition types. However, similarly to what was observed in previous sections, \textit{O-Ideal}, \textit{O-Curriculum}, \textit{RL-ALPEXP} and \textit{RL-LogP} appear to be the best performing Experimenters. More importantly, these results show a significant improvement over the performance without any hypotheses.

\subsection{Generating hypotheses with WorldLLM vs fine-tuning}
WorldLLM improves an LLM's forward modeling abilities by providing it natural language hypotheses about the world. In this section, we study how our approach compares to a more classic gradient-based approach where the LLM is directly fine-tuned on collected transitions. In addition to avoiding gradient computation through the LLM and being human interpretable, the theories produced by WorldLLM are expected to better generalize to changes in the environment.  

We use the transitions collected by RL-LogP and fine-tune our Statistician's LLM on them using a causal language modeling objective (i.e., we maximize the likelihood of the next state's tokens given the state-action pair). We name this new approach \textit{F-LogP}. When evaluated on its top-3 constrained decoding accuracy over the test set as in Section~\ref{sec:worldllm-eval_world_model}, F-LogP obtains perfect results (i.e., accuracy of $1$) over all transition types.  In the following sections, we study the generalization abilities of both methods and provide a deeper analysis of their impact on log-likelihoods.

\paragraph{Generalization to changes in the environment} \label{sec:worldllm-generalization}
We now assess whether using natural language hypotheses in the prompt produces better generalization abilities for the LLM-based world model than fine-tuning it on collected transitions. For this, we create a generalization environment in which the syntax of observations is changed (see Figure~\ref{prompt:generalization}). We use this new environment to report the top-3 constrained decoding accuracy on our test set, as in previous sections.

\begin{table}
\centering
{\tiny
\begin{tabular}{lcccccc}
\toprule
\textbf{Methods} & \textbf{Standing} & \textbf{Holding 1} & \textbf{Holding 2} & \textbf{Grow Plant} & \textbf{Grow S.H.} & \textbf{Grow B.H.} \\
\midrule
No Hypo. & 0.28 $\pm$ 0.00 & \textbf{0.85 $\pm$ 0.00} & 0.00 $\pm$ 0.00 & 0.00 $\pm$ 0.00 & 0.00 $\pm$ 0.00 & 0.00 $\pm$ 0.00 \\
\midrule
O-Random & 0.69 $\pm$ 0.16 & 0.45 $\pm$ 0.26 & 0.00 $\pm$ 0.00 & 0.00 $\pm$ 0.00 & 0.00 $\pm$ 0.00 & 0.00 $\pm$ 0.00 \\
O-Ideal & 0.73 $\pm$ 0.23 & 0.46 $\pm$ 0.19 & 0.00 $\pm$ 0.00 & 0.00 $\pm$ 0.00 & 0.00 $\pm$ 0.00 & 0.00 $\pm$ 0.00 \\
O-Hardest & 0.33 $\pm$ 0.33 & 0.71 $\pm$ 0.33 & 0.00 $\pm$ 0.00 & 0.00 $\pm$ 0.00 & 0.00 $\pm$ 0.00 & 0.00 $\pm$ 0.00 \\
O-Curriculum & 0.46 $\pm$ 0.34 & 0.72 $\pm$ 0.29 & 0.00 $\pm$ 0.00 & 0.00 $\pm$ 0.00 & 0.00 $\pm$ 0.00 & 0.00 $\pm$ 0.00 \\
\midrule
RL-ALPEXP & 0.65 $\pm$ 0.16 & 0.42 $\pm$ 0.30 & 0.00 $\pm$ 0.00 & 0.00 $\pm$ 0.00 & 0.00 $\pm$ 0.00 & 0.00 $\pm$ 0.00 \\
\midrule
RL-LogP & 0.43 $\pm$ 0.16 & 0.30 $\pm$ 0.23 & 0.00 $\pm$ 0.00 & 0.00 $\pm$ 0.00 & 0.00 $\pm$ 0.00 & 0.00 $\pm$ 0.00 \\
RL-ALP & 0.76 $\pm$ 0.18 & 0.61 $\pm$ 0.31 & 0.00 $\pm$ 0.00 & 0.00 $\pm$ 0.00 & 0.00 $\pm$ 0.00 & 0.00 $\pm$ 0.00 \\
\midrule
F-LogP & \textbf{0.88 $\pm$ 0.17} & 0.24 $\pm$ 0.25 & 0.00 $\pm$ 0.00 & 0.07 $\pm$ 0.13 & \textbf{0.21 $\pm$ 0.32} & \textbf{0.67 $\pm$ 0.44} \\
\bottomrule
\end{tabular}}
\caption{Top 3 performance on Constrained Decoding for each algorithm by transition type on the test set using the generalization environment. We also report the performance without any hypotheses given (No Hypo.).}
\label{tab:4-worldllm-constrained-decoding-finetuning-generalization}
\end{table}

Results from Table~\ref{tab:4-worldllm-constrained-decoding-finetuning-generalization} show an important deterioration of the performance for all methods. F-LogP manages to maintain a non-zero accuracy on complex transition types while WorldLLM-based world models do not perform better than the Statistician without any hypotheses. These results hint that WorldLLM may struggle in finding abstract hypotheses that generalize. The next section provides additional insights on these results.

\begin{figure}[hbt]
    \centering
    \subfloat[No changes to the environment.]{\includegraphics[width=.5\textwidth]{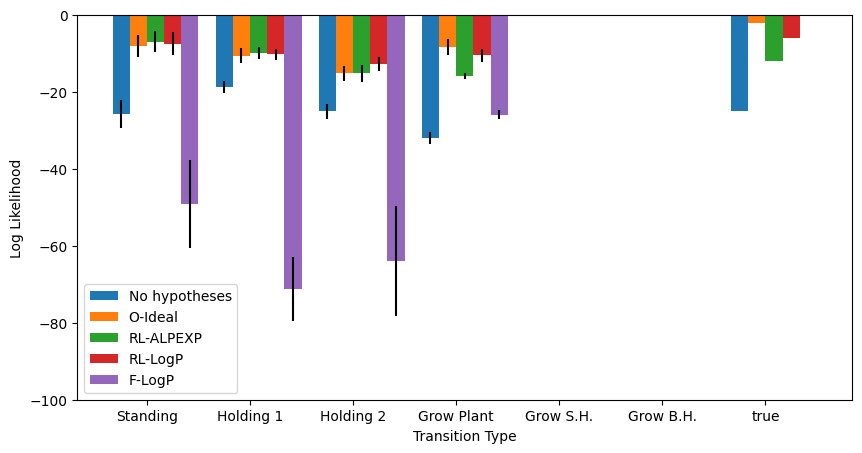}}
    \hfill
    \subfloat[Generalization environment.]{\includegraphics[width=.5\textwidth]{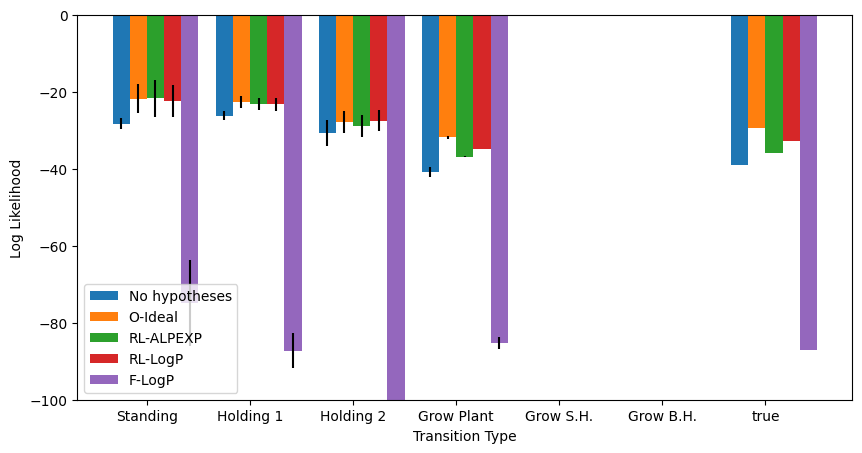}}
    \caption{Log-likelihoods (averaged by transition type) produced by the Statistician when facing an instance of a \textit{Grow Plant} transition. We show how using the generalization environment from Section~\ref{sec:worldllm-generalization} affects the log-likelihoods (b).}
    \label{fig:4-worldllm-DistribLogPGeneralization}
\end{figure}

\paragraph{Effect on log-likelihoods}
We now propose to study how both methods affect the Statistician's log-likelihoods. We show in Figure~\ref{fig:4-worldllm-DistribLogPGeneralization} the log-likelihoods (averaged by transition type) produced by the Statistician when facing an instance of a \textit{Grow Plant} transition. 

Looking at results when using the classic environment (a), WorldLLM increases the log-likelihood of all the possible transitions (no matter the Experimenter) while maintaining the correct transition more likely. In comparison, the causal language modeling objective used for F-LogP increases the log-likelihood of the correct transition while significantly decreasing the log-likelihood of all the other transitions. These results illustrate the difference in the two optimization processes: while fine-tuning aims to increase the distance in log-likelihoods between the correct transition and the other ones, WorldLLM focuses on increasing the log-likelihood of the correct transition, regardless of the others.

When focusing on performance with the generalization environment, WorldLLM leads to very few changes compared to when no hypotheses are given. On its side, F-LogP produces very low log-likelihoods for all transitions and keeps the correct one slightly above the others. Overall, these results show that both approaches generalize poorly.

One explanation for WorldLLM's poor generalization performance lies in the hypotheses produced. We provided samples in Appendix~\ref{app:generated_hypotheses} which show that our Scientist notably used examples of transitions in the hypotheses (even though some hypotheses did mention abstract categories or even mentioned objects not in our environment). Such examples no longer align with the generalization environment's syntax. Multiple explanations can be given for this. First, our current experiments rely on a Scientist's LLM with limited capabilities. Second, while our objective in \ref{eq:4-worldllm-objective} minimizes the number of hypotheses, the current implementation of WorldLLM only focuses on maximizing the Statistician's log-likelihood. We argue that favoring short hypotheses could help abstract theories emerge. We leave these investigations for future work.

\section{Related work}
\subsection{Building forward models}
Building forward models is a long-established research direction \citep{ha_world_2018, moerland_model-based_2022}. In addition to producing more interpretable agents, forward models were particularly used in RL to improve sample efficiency, allowing the agent to train on imagined transitions or perform multi-step planning \citep{hafner_dream_2020, hafner_mastering_2022, hafner_mastering_2024, schrittwieser_mastering_2020}. Parallel to this, learning a forward model has been used as an intrinsic reward to foster RL policies' exploration \citep{schmidhuber_possibility_1991, pathak_curiosity-driven_2017, burda_large-scale_2018, oudeyer_what_2007}. WorldLLM employs a similar curiosity-driven scheme where our RL-based Experimenters use reward signals from prior work to promote exploration and collect transitions that help improve the forward model. However, our method differs in the way the model is improved: instead of using a gradient-based approach to fine-tune the LLM-based forward model, as done in \citep{xiang_language_2023}, WorldLLM searches for natural language theories that improve the forward model when given in the prompt. While recent work studied how a forward model could leverage external textual information such as a manual explaining the environment \citep{zhang_language-guided_2024, dainese_reader_2023}, our approach iteratively generates and refines its own explanations. 

\subsection{Inductive reasoning for world modeling} 
The importance of theories in shaping one's representations about the world was argued as central, in both children development and adults \citep{piaget_construction_1954,gopnik_scientist_1996}
Building on this, the literature of theory-based RL has been advocating for augmenting artificial agents with inductive reasoning \citep{tsividis_human-level_2021}. In particular, \citep{tsividis_human-level_2021, tang_worldcoder_2024, li_automated_2024} proposed approaches based on LLMs that generate programs representing the theories about an environment's dynamics. The use of programs allowed the authors to execute and compare how the proposed theories match empirical evidence. Parallel to this, \citep{piriyakulkij_doing_2024, wang_hypothesis_2023} proposed to generate theories using natural language. While the former constrained the generated theories to programs that can be easily evaluated in their environment, the latter used the generated theories to condition a program generator to simulate the environment's dynamics. WorldLLM also builds on the theory-based RL literature and closely relates to \citep{piriyakulkij_doing_2024}. Notably, WorldLLM also leverages Bayesian inference with an LLM as the proposal distribution to generate the theories. However, the major difference lies in how the produced theories are used: while \citep{piriyakulkij_doing_2024} constrain the format of theories such that they can be directly tested in the environment, our work focuses on giving these theories in another LLM's prompt to improve its forward modeling capabilities. \textit{In other words, our work strongly differs in its perspective of producing theories that improve our LLM's world modeling abilities instead of producing correct theories}. Nonetheless, the Bayesian inference framework we used in our Scientist is largely inspired by \citep{piriyakulkij_doing_2024}. Both our work and theirs also consider an active learning paradigm to choose which experiments to perform for theories' refinement. While their approach computes exact Information Gain of all possible experiments, ours relies on curiosity-driven RL agents and Information Gain approximates as rewards to overcome the intractability of exact Information Gain in most environments.

\section{Discussion}
In this paper, we introduced WorldLLM, a framework for enhancing world modeling abilities in LLMs. WorldLLM generates natural language hypotheses—interpretable, theory-like statements—which condition the LLM when predicting state-action outcomes. This approach enables world modeling without expensive fine-tuning of the LLM itself and avoids the need for a handcrafted theory space. Our experiments demonstrated that the adversarial dynamic between a Scientist (hypothesis generator) and simple curiosity-driven Experimenters (data collectors) supports broad exploration and the discovery of environment dynamics.

Despite promising results, our initial implementation of WorldLLM presents several limitations. First, the Playground-Text environment is structurally simple, and the natural ordering created by the most challenging transitions being longer and harder to predict is not expected in more complex environments—potentially reducing the effectiveness of RL-LogP. Evaluating WorldLLM in more complex environments is therefore a critical direction for future work \citep{ying_assessing_2025}.

Second, our current Experimenters use lightweight neural networks and symbolic inputs. A more compelling alternative would be to integrate LLM-based Experimenters, possibly unifying all WorldLLM components with a single GLAM-equipped LLM that learns both to control and to improve its forward model via online RL. Furthermore, our experiments only used one LLM model, which showed limitations in the Scientist role—particularly in generating abstract, high-level hypotheses. 

In addition to scaling the LLM, we argue that the emergence of such abstract hypotheses may result from other modifications to our implementation. A key advantage of Bayesian inference in this context is the ability to incorporate \textit{priors}, which constrain the theory space. While a length-based prior might encourage abstract hypotheses, we believe richer priors—such as “error maps” that represent structured patterns of predictive failure, as described by \citet{schulz_chapter_2012}—may be necessary to better replicate human-like theory refinement.

Additionally, our current Metropolis-based approach for updating hypotheses only considers the most recent transitions, which can lead to forgetting, especially with non-stationary Experimenters. This was primarily motivated by practical constraints: the LLM's limited context window and the cost of evaluating likelihoods on large transition sets. These issues could be mitigated by storing past transitions in a replay buffer or employing more sophisticated Bayesian inference methods (e.g., particle filters), which would also allow tracking multiple particles to avoid local optima—a current limitation evidenced in Appendix~\ref{app:additional_results}.

Finally, our exploration strategy relied solely on simple curiosity-driven signals focused on poorly predicted transitions. Other intrinsic motivation mechanisms have been studied in both humans and artificial agents. A particularly important one is \textit{autotelic learning}, where learners generate, select, and attempt to master self-defined goals \citep{steels_autotelic_2004, colas_autotelic_2022}. Studying the use of such more advanced curiosity signals could significantly improve the quality of collected transitions.

In conclusion, WorldLLM represents a significant advancement, opening up promising new avenues for developing more grounded LLMs that can efficiently predict and explain their operational environments. While the current iteration has certain limitations, these also present opportunities for substantial enhancements and future innovations in the methodology.

\subsubsection*{Acknowledgments}
This work was granted access to the HPC resources of IDRIS under the allocation A0171011996 made by GENCI.

\newpage
\bibliographystyle{Styles/iclr2025/iclr2025_conference}
\bibliography{main}

\newpage
\appendix

\section{The Playground-Text Environment}
\label{app:experimental_setup}
We use the Playground-Text environment from \citep{gaven2024sacglamimprovingonlinerl}, itself being a textual adaptation of the Playground environment introduced by \citep{colas_language_2020}. This textual environment returns, at each time step, a description of its state and possible actions. In our experiments, we used a modified version of Playground-Text where the current state still describes the whole environment's state, but the next state only describes how the performed action changed the environment (see Figure \ref{fig:PlaygroundState}). 

Playground-Text features 4 different classes of objects with different varieties: \textit{Water}, \textit{Plants} (carrot, potato, beet, berry, and pea), \textit{Small Herbivores} (pig, cow, and sheep), and \textit{Big Herbivores} (elephant, giraffe, rhinoceros). As depicted in Figure \ref{fig:4-worldllm-playground_env}, all objects except \textit{Water} possess two states: young (seed for plants and baby for animals) and grown. These objects transition from their young state to their grown one by consuming other objects.

Three different types of actions are possible: \textit{Go to <object>}, to move to the selected object, \textit{Grasp}, to put the object the agent is standing on in its inventory (which has two slots) and \textit{Release <object>/all} to release one or all objects from the inventory on the object the agent is currently standing on. The resulting transitions can be categorized into 6 different types: \textit{Standing}, \textit{Holding 1}, \textit{Holding 2}, \textit{Grow Plant}, \textit{Grow Small Herbivore} or  \textit{Grow Big Herbivore}. Growing transitions are triggered when the appropriate objects are released on a young one (see the technology tree in Figure~\ref{fig:4-worldllm-playground_env}).   
All actions involving non-existent objects in the current scene, as well as release actions that do not result in a transformation (e.g. releasing water onto water) are considered illegal and masked.

\begin{figure}[hbt]
\begin{promptbox}{}
\textit{State}:
You see the baby cow, the water, the potato seed, the baby rhinoceros, the water, the pea seed, the water, the potato seed. You are next to the potato seed. In your inventory, there is nothing. \\
\textit{Action}: 
You go to the water. \\
\textit{Change}: 
You are standing on the water. \\ \\
\textit{State}:
You see the baby cow, the water, the potato seed, the baby rhinoceros, the water, the pea seed, the water, the potato seed. You are next to the water. In your inventory, there is nothing. \\
\textit{Action}: 
You pick up the object. \\
\textit{Change}: 
In your inventory, there is the water. \\ \\
\textit{State}:
You see the baby cow, the potato seed, the baby rhinoceros, the water, the pea seed, the water, the potato seed. In your inventory, there is the water.\\
\textit{Action}: 
You go to the potato seed. \\
\textit{Change}: 
You are standing on the potato seed. \\ \\
\textit{State}:
You see the baby cow, the potato seed, the baby rhinoceros, the water, the pea seed, the water, the potato seed. You are next to the potato seed. In your inventory, there is the water. \\
\textit{Action}: 
You give the water. \\
\textit{Change}: 
The objects transform into the potato. \\
\end{promptbox}
\caption{An example of a trajectory in Playground-Text.}
\label{fig:PlaygroundState}
\end{figure}

\section{WorldLLM details} \label{app:worldllm}
In this section, we provide details about our WorldLLM implementation. All our LLMs were loaded and prompted using the \textit{transformers} library from Hugging Face. We also provide our code for reproduction on the following repository: \url{https://github.com/flowersteam/WorldLLM}.

\subsection{Statistician}
The LLM used for the Statistician is \textit{Phi-3-mini-4k-instruct} quantized to 4 bits. The prompt used is shown in Figure~\ref{prompt:Statistician}.

\begin{figure}[hbt]
    \begin{promptbox}{}
        \textbf{System prompt:}\\
        You like doing a lot of puzzles. Please answer with a brief answer and be as precise as you can. \\
        \textbf{User prompt:}\\
        You are in an environment that contains multiple objects. It can contain water, plant seeds(carrot, porato, beet, berry and pea seeds), small herbivores(pig, cow and ship) and large herbivores(elephant, giraffe, rhinoceros). You can move objects, like water, plants or herbivores and release them on another object to make them interact and transform into a new object. You know that: \\ <hypothesis> \\
        Your objective is to predict the next change in the environment given the state of the environment and the action taken. \\
        The last state was: 
        <state>  \\
        The last action was: 
        <action>\\
        The change is: 
        \textbf{Assistant prompt:}\\
        <change>
    \end{promptbox}
    \caption{Prompt used for the Statistician. The <hypothesis> corresponds to the hypotheses to test. The <state>, <action> and <change> represent the current state, the action taken, and the resulting change, respectively, for the given transition.}
    \label{prompt:Statistician}
\end{figure}

\subsection{Scientist}
The LLM used for the Scientist is the same as the one used for the Statistician. We show its prompt in Figure~\ref{prompt:Scientist}.

\begin{figure}[hbt]
    \begin{promptbox}{}
        \textbf{System prompt:}\\
        You like doing a lot of puzzles. Please answer with a brief answer and be as precise as you can. \\
        \textbf{User prompt:}\\
        You are in an environment that contains multiple objects. It can contain water, plant seeds(carrot, potato, beet, berry and pea seeds), small herbivores(pig, cow and ship) and large herbivores(elephant, giraffe, rhinoceros). You can move objects, like water, plants or herbivores and release them on another object to make them interact and transform into a new object. Your previous experiences were: \\ <trajectories> \\ Can you find a set of easily understandable and concise hypotheses to predict how the environment will change based on these trajectories? You can take inspiration from the previous rules:\\
        <previous hypothesis> \\
        You also know that the previous set of rules failed the most on those trajectories: \\
        <worst trajectories>
        \\
        Answer with just the hypothesis.
    \end{promptbox}
    \caption{Prompt used for the Scientist. <trajectories> correspond to the trajectories collected by the Experimenter, <previous hypothesis> corresponds to the set of hypotheses that have been accepted at the last iteration and <worst trajectories> to the trajectories where the previous hypotheses obtained the worst log-likelihood.}
    \label{prompt:Scientist}
\end{figure}

\subsection{Experimenter}
For the Experimenter, we provide further details for both our oracles and RL-based policies.

\subsubsection{Oracles} \label{app:heuristic-agents}
\begin{figure}
    \centering
    \includegraphics[width=0.49\linewidth]{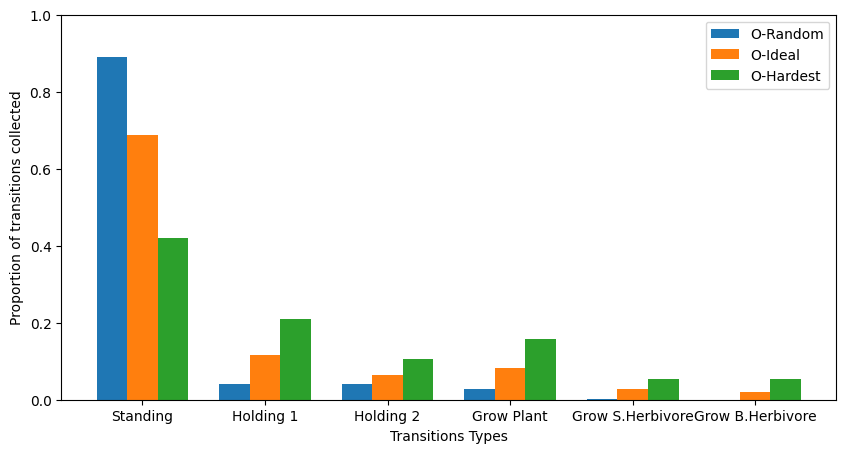}
    \includegraphics[width=0.49\linewidth]{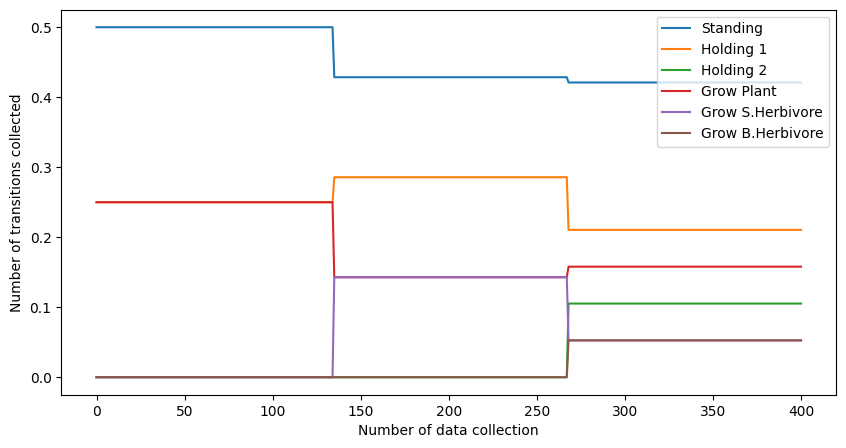}
    \caption{Proportion of collected transition types for each oracle. We show on the left the distribution of our stationary oracles and on the right the evolving distribution of O-Curriculum.}
    \label{fig:proportion-transition-oracle}
\end{figure}

We show in Figure~\ref{fig:proportion-transition-oracle} the distribution of transitions collected by each of our four oracles.

\subsubsection{RL agents}\label{app:rl-agents}
For our RL agents we used Masked PPO from stable-baselines3 contributions\footnote{\url{https://sb3-contrib.readthedocs.io/en/master/modules/ppo_mask.html}}. To accelerate training, we used 15 parallel environments and collected 8 trajectories per environment. We show the hyperparameters used in Table~\ref{tab:hyperparameters RL}.

Concerning the reward signals, we provide in Algorithm~\ref{alg:worldllm-rl-alpexp} the details on how RL-ALPEXP compute its ALP reward. RL-ALPEXP introduces an additional hyperparameter $\alpha$ which regulates the reward's smoothing of the reward. For each transition type, $\alpha$ corresponds to the weight of the previous rewards obtained for this transition type. Its role is highlighted in our study from Appendix~\ref{app:rl-alp-failed} but $\alpha$ can be interpreted as an adjustment factor determining the extent to which rewards from previous steps are retained. A value of $0$ for $\alpha$ implies that only the reward from the most recent step is considered, which corresponds to RL-ALP, while a value of $1$ indicates that the rewards are accumulated from the beginning. After a manual hyperparameter search (see Appendix~\ref{app:rl-alpexp_hyperparams}), we chose $\alpha=0.9$ across all transition types.

\begin{table}[hbt]
    \centering
    \begin{tabular}{|c|c|}
        \hline
        \textbf{hyperparameter} & \textbf{value} \\
        \hline
        gamma & $0.99$ \\
        lr & $5e-4$ \\
        vf coef & $0.5$\\
        gae $\lambda$ & $0.95$\\
        entropy & $0.01$ \\
        epochs & 20 \\
        architecture & $64\times64$\\
        \hline
    \end{tabular}
    \caption{PPO hyperparameters used for our RL agents}
    \label{tab:hyperparameters RL}
\end{table}

\begin{algorithm}
\caption{RL-ALPEXP}
\begin{algorithmic}[1]
\State \textbf{Input:} $\alpha$ moving average coefficient, $\mathcal{I}$ set of transition types, $T$ total number of iterations, $n_{steps}$ number of Metropolis steps at each iteration, $f$ function returning the type of a transition given a transition, $\pi$ the Experimenter
\State \textbf{Initialize} $(m_i)_{i \in \mathcal{I}} \gets 0$ \Comment{Initialize moving average for each type}
\For{$i=0$ to $n$}
    \State Collect trajectories $\mathcal{D}_t \leftarrow$env$(\pi)$
    \State Perform $n_{steps}$ Metropolis steps
    \For{$\tau \in \mathcal{D}_t$}
        \State Compute reward $(r_{alp})_\tau$  
        \State $(r_{alpexp})_\tau=\alpha*m_{f(\tau)}+(1-\alpha)*(r_{alp})_\tau$\Comment{Compute ALPEXP reward}
    \EndFor
    \For{$i \in \mathcal{I}$}
        \State $\tilde{m}_\beta = \frac{\sum\limits_{\tau \in \mathcal{D}_t}  \mathbf{1}_{f(\tau)=\beta}(r_{alp})_\tau}{\sum\limits_{\tau \in \mathcal{D}_t}  \mathbf{1}_{f(\tau)=\beta}}$\Comment{Compute mean over transitions}
        \State $m_\beta=\alpha*m_\beta + (1-\alpha)*\tilde{m}_\beta$\Comment{Update moving average}
    \EndFor
    \State Update Experimenter $\pi$ 
\EndFor
\end{algorithmic}
\label{alg:worldllm-rl-alpexp}
\end{algorithm}

\section{Generalization Tests}
We show in Figure~\ref{prompt:generalization} how the syntax of observations was changed in our generalization tests. In particular, we modified the words associated to the transition (e.g. standing, transform) and inverted the sentence. These changes were also applied to the prompts used by the Scientist and the Statistician.

\begin{figure}[hbt]
    \begin{promptbox}{}
        \textbf{Standing: }
        You are standing on the \textit{<object>}. $\longrightarrow$ The \textit{<object>} is beneath you.
        \\
        \textbf{Holding 1: }
        In your inventory, there is the \textit{<object>}. $\longrightarrow$
        The \textit{<object>} is in your grasp.
        \\
        \textbf{Holding 2: }
        In your inventory, there are the \textit{<object>} and the \textit{<object>}. $\longrightarrow$
        The \textit{<object>} and the \textit{<object>} are in your grasp.
        \\
        \textbf{Grow Plant: }
        The objects transform into the \textit{<object>}. $\longrightarrow$
        The \textit{<object>} results from combining the objects.
        \\
        \textbf{Grow Small Herbivore: }
        The objects transform into the \textit{<object>}. $\longrightarrow$
        The \textit{<object>} results from combining the objects.
        \\
        \textbf{Grow Big Herbivore: }
        The objects transform into the \textit{<object>}. $\longrightarrow$
        The \textit{<object>} results from combining the objects.
    \end{promptbox}
    \caption{Changes in the observations for the generalization environment.}
    \label{prompt:generalization}
\end{figure}

\section{Additional Results} \label{app:additional_results}
In this section, we first provide the complete evolution of log-likelihoods on our test set from Section~\ref{sec:worldllm-rl}. Then, we provide additional studies and experiments to improve one's comprehension of WorldLLM's behavior. These investigations notably explore the reasons behind the failure of RL-ALP and its differences with RL-ALPEXP.

\subsection{Evolution of likelihoods}
\begin{figure}
    \centering
    \includegraphics[width=0.8\linewidth]{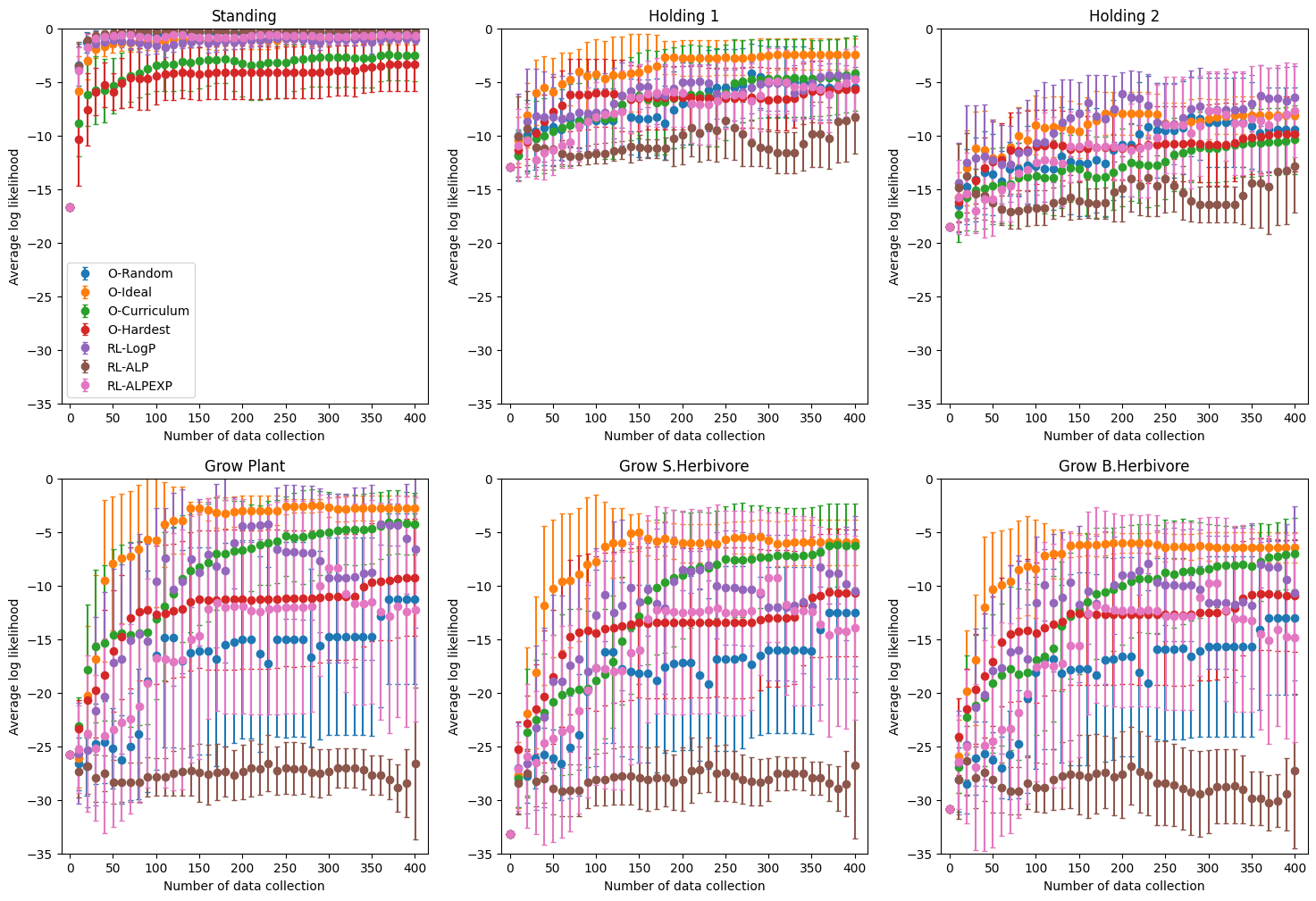}
    \caption{Evolution of log-likelihoods on the test set for all methods.}
    \label{fig:loglikelihoods_all}
\end{figure}

We report in Figure~\ref{fig:loglikelihoods_all} the evolution of log-likelihoods on our test set for all the evaluated methods.

\subsection{Analysis of the Metropolis algorithm in the Scientist} 
\subsubsection{Hypotheses Retention Rate}\label{app:RetentionRate}
\begin{figure}[hbt]
    \centering
    \includegraphics[width=0.8\linewidth]{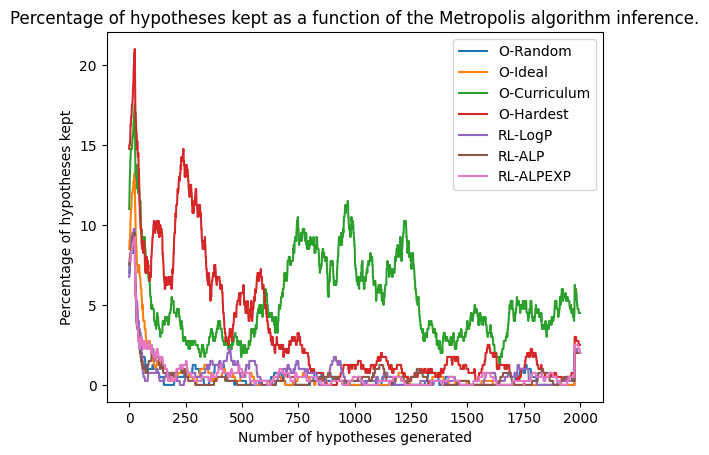}
    \captionof{figure}{Evolution of the hypotheses retention rate for the different Experimenters.}
    \label{fig:proportion-hypotheses-accepted}
\end{figure}

First, we analyze how the Metropolis algorithm performs when using an LLM as both the proposal and target distribution. In particular, we study the evolution of accepted hypotheses throughout WorldLLM's iterations when using different Experimenters (Figure~\ref{fig:proportion-hypotheses-accepted}).

Our results show a high percentage of accepted hypotheses (around 20\%) at the beginning of the experiment. However, as iterations progress, it becomes increasingly challenging for the Scientist to generate more effective hypotheses, leading to a decline in the retention rate (it reaches 2\%). As highlighted in our manuscript, a developmental strategy is key in WorldLLM. This is here illustrated by O-Curriculum's ability to maintain a higher retention rate than all other oracles. Additionally, preliminary experiments performed with Phi4 suggested that employing a larger LLM for the Scientist improves hypotheses generation and leads to higher retention rates for the same Statistician model.

\subsubsection{Embeddings of the hypotheses} \label{app:embeddings}
In order to compare the hypotheses generated by our Scientist, we used a sentence transformer, \textbf{all-mpnet-base-v2}\footnote{\url{https://huggingface.co/sentence-transformers/all-mpnet-base-v2}}, to project hypotheses in a latent space. We then used T-SNE\citep{maaten_visualizing_2008} to obtain a 2D representation of the embeddings. As the number of generated hypotheses is high, we plotted only 1/10 of them for each Experimenter and seed. We show these embeddings in Figure~\ref{fig:embeddings}. 

The results show that O-Curriculum and O-Hardest mostly produce similar hypotheses. This indicates a Scientist's tendency to refine previous hypotheses rather than generating brand new ones. This corroborates our results on retention rate from previous section. 
We also compute an inertia score across all seeds for each Experimenter:
\begin{equation}
    \forall \beta \in \mathcal{M}, \quad I =\sum\limits_{i=1}^{n} (x_i -C_\beta)^{2} \qquad \text{with} \quad C_\beta = \sum\limits_{i=1}^{n} \frac{x_i}{n} 
\end{equation}
with $\mathcal{M}$ the set of Experimenters, $x_i$ the 2D projection from T-SNE for each set of hypotheses and $n$ the total number of hypotheses. 

Results indicate that the clusters produced by O-Curriculum and O-Hardest are each produced by a different seed. This highlights the current limitation of our Bayesian inference approach where a single particle is used and never reset. As a result, our Scientist may easily get stuck in local optima.

\begin{minipage}{0.5\textwidth}
    \centering
    \includegraphics[width=\linewidth]{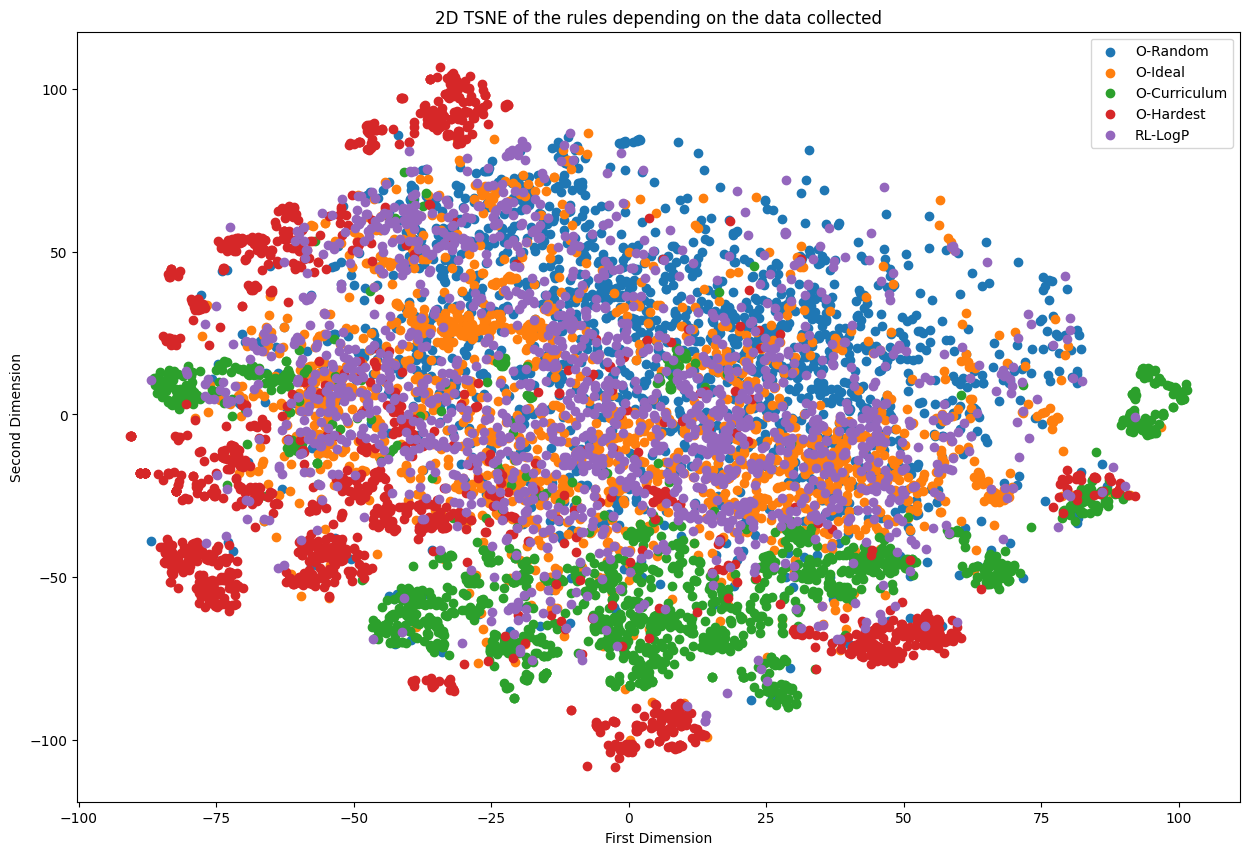}
    \captionof{figure}{T-SNE embeddings obtained using the  \textit{all-mpnet-base-v2} sentence transformer on the hypotheses generated by different Experimenters.}
    \label{fig:embeddings}
\end{minipage}
\begin{minipage}{0.5\textwidth}
    \centering
    \begin{tabular}{|c|c|}
        \hline
        \textbf{Methods} & \textbf{Inertia Score} \\
        \hline
        O-Random     & 4.12e6 \\
        O-Ideal      & 4.38e6 \\
        O-Curriculum & 2.77e6 \\
        O-Hardest    & 2.88e6 \\
        RL-LogP      & 4.52e6 \\
        RL-ALP       & 3.62e6 \\
        RL-ALPEXP    & 4.92e6 \\
        \hline
    \end{tabular}
    \label{tab:embedding-cluster-inertia}
    \captionof{table}{Mean of the inertia score on the T-SNE embeddings across the different seeds for each Experimenter.}
\end{minipage}

\subsection{Analysis of generated hypotheses} \label{app:generated_hypotheses}
In this section, we propose a qualitative study of hypotheses produced by WorldLLM when using O-Curriculum, O-Ideal and RL-LogP as Experimenters (see Figure~\ref{fig:hypotheses_examples}).

One can observe that each set of hypotheses covers multiple transition types. This can be explicit, as seen in the hypotheses from O-Ideal, or more abstract, as the ones from O-Curriculum, which replace specific animals and objects with abstract representations. A closer examination of O-Curriculum's hypotheses shows that new entities that do not exist in Playground-Text such as "bear", "mouse" or "wolf" are introduced. This aligns with our argument that natural language hypotheses could help the world model generalize.
However, our poor generalization results can be explained by the fact that most hypotheses contain examples of transitions from the environment. Such examples tremendously help our Statistician predict outcomes but also significantly impair its generalization capabilities.

\begin{figure}[hbt]
\begin{promptbox}{}
    \footnotesize{
    \textbf{O-Curriculum: } \\
    1. If the action is 'You go to' and the object is a seed or gentle creature (cow, elephant, sheep, giraffe, bear, mouse, wolf), predict "You are standing on the [object]."\\2. If you are standing on an animal's seed and the action is 'You give the water', predict "The objects transform into the [animal]."\\3. If the object is a seed or gentle creature (cow, elephant, sheep, giraffe, bear, mouse, or wolf) and the action is 'You give the water', predict "The objects transform into the [animal]."\\4. If you are standing on the water and the action is 'You pick up', predict "In your inventory, there is the water."\\5. If you are standing on the water and the action is 'You pick up the object', predict "In your inventory'\\
    \textbf{O-Ideal: } \\
    1. The action was: You go to the water, and the change was: You are standing on the water.\\2. The action was: You pick up the object, and the change was: In your inventory, there is the water.\\3. The action was: You go to the beet seed, and the change was: You are standing on the beet seed.\\4. The action was: You give the water, and the change was: The objects transform into the beet.\\5. The action was: You pick up the object, and the change was: In your inventory, there is the beet.\\6. The action was: You go to the baby cow, and the change was: You are standing on the baby cow.\\7. The action was: You give the beet, and the change was: The objects transform into the cow.\\8. The action was: You go to the pe' \\
    \textbf{RL-LogP: } \\
    1. Go to the water.\\Change: You are standing on the water.\\2. Pick up the object.\\Change: In your inventory, there is the water.\\3. Go to the water.\\Change: You are standing on the water.\\4. Pick up the object.\\Change: In your inventory, there are the water and the water.\\5. Go to the berry seed.\\Change: You are standing on the berry seed.\\6. Give the water.\\Change: The objects transform into the berry.\\7. Go to the water.\\Change: You are standing on the water.\\8. Pick up the object.\\Change: In your inventory, there are the water and the water.\\9. Go to the pea seed.\\Change: You are standing on the pea seed.\\10. Give the water.\\Change: The objects transform into
    }
\end{promptbox}
\caption{Examples of accepted hypotheses.}
\label{fig:hypotheses_examples}
\end{figure}

\subsection{A deeper study RL-ALP}
In this section, we provide a deeper analysis of RL-ALP's failure. While expected to be more robust than RL-LogP (e.g. to the "noisy TV" problem), our results showed RL-ALP was unable to collect other transitions than \textit{Standing}, leading to very poor hypotheses. We begin this section by studying how directly using the per-transition ALP as reward for an RL policy is challenging. Then, show how RL-ALPEXP differs and helps stabilize training.

\subsubsection{Failure of RL-ALP as the reward}\label{app:rl-alp-failed}
\begin{figure}
    \centering
    \includegraphics[width=0.5\linewidth]{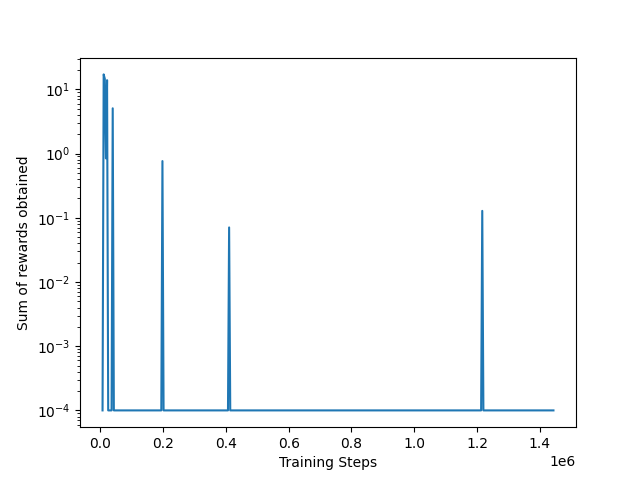}
    \caption{The evolution of rewards obtained by the RL agent when using RL-ALP for a single seed. We plot the sum of rewards obtained for each PPO update.}
    \label{fig:rewards-alp}
\end{figure}

To get a better grasp of the RL policy's behavior when using RL-ALP, we plot the evolution of rewards obtained by the policy in Figure~\ref{fig:rewards-alp}. We observe that most rewards equal $0$. Higher rewards are only obtained in the first data collections when \textit{Standing} transitions are not well predicted by the Statistician. After gathering such transitions, the policy eventually converges to a random one. This absence of rewards can be explained by our reward definition from Eq.~\ref{eq:4-worldllm-RL-ALP}.

A positive reward is obtained only if a set of hypotheses are accepted between the previous and the current iteration. However, as discussed in Appendix~\ref{app:RetentionRate}, the number of accepted hypotheses tends to quickly drop and eventually reach a very low percentage. As a result, very few improvements (or degradation) are made to the Statistician's log-likelihoods, leading to zero rewards.

\subsubsection{Ablation study on RL-ALPEXP} \label{app:rl-alpexp_hyperparams}
In Appendix~\ref{app:rl-agents}, we detailed RL-ALPEXP, which includes a new hyperparameter $\alpha$. This hyperparameter governs the weighting of smoothing applied to the reward computed on various transition types over time. In this section, we explore the impact of this hyperparameter.

\begin{figure}
    \centering    \includegraphics[width=0.24\linewidth]{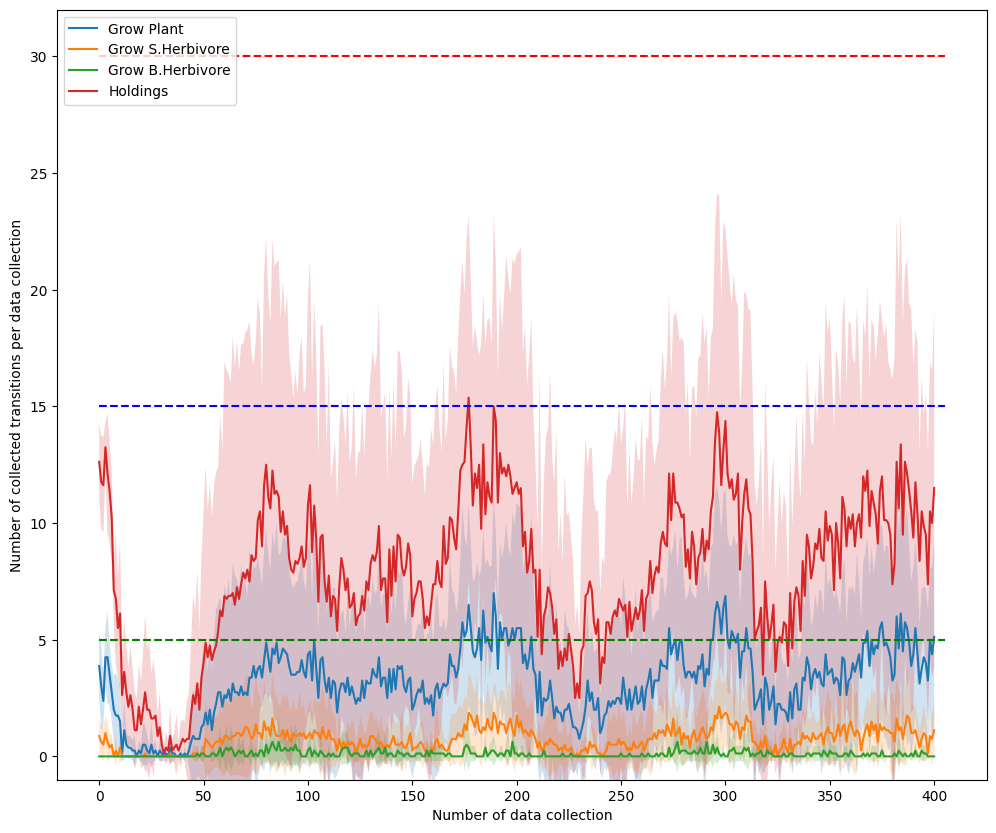}    \includegraphics[width=0.24\linewidth]{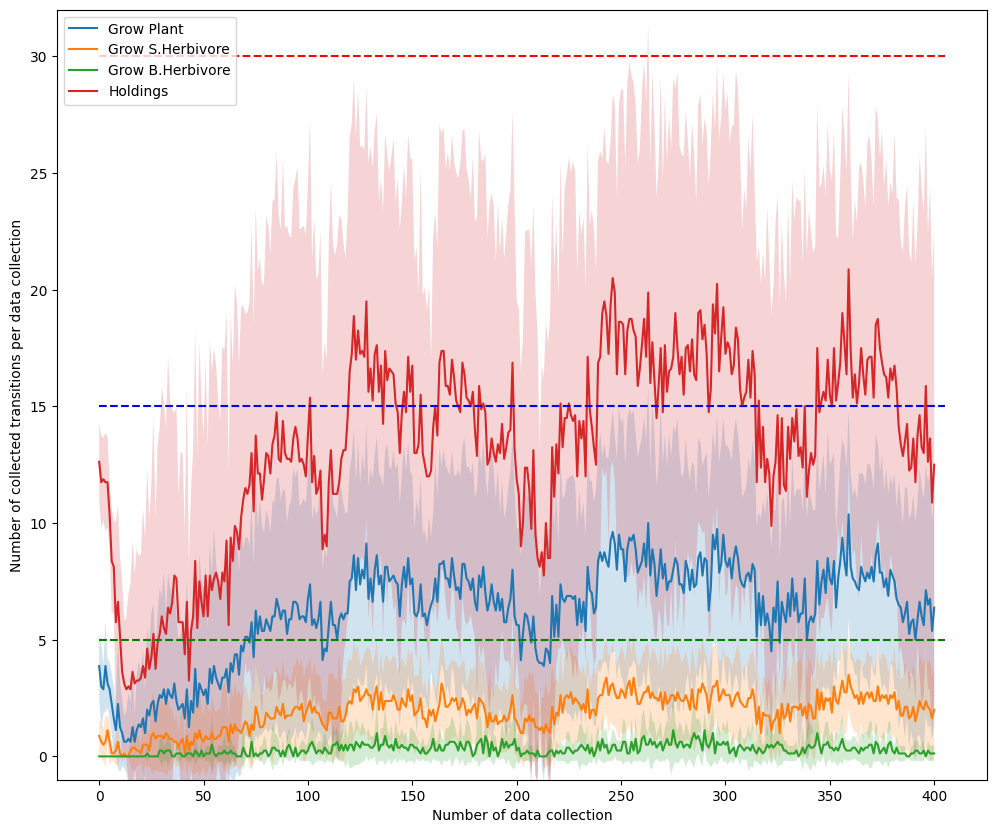}    \includegraphics[width=0.24\linewidth]{Images/New/evolution-proportion-transition_alpexp09.png}    \includegraphics[width=0.24\linewidth]{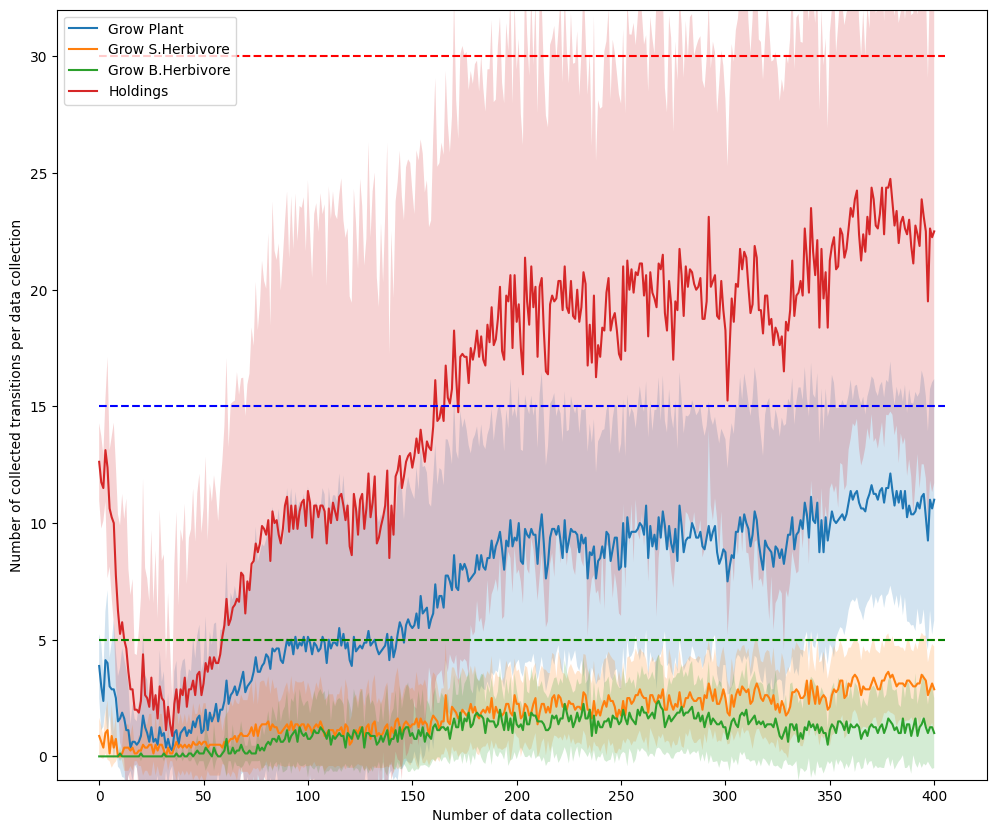}
    \caption{Transitions collected by the Experimenter when using RL-ALPEXP for $\alpha=\{0.6,0.75,0.9,0.95\}$ respectively.}
    \label{fig:rl-alpexp-alpha}
\end{figure}

To assess how $\alpha$ influences RL-ALPEXP, we conducted qualitative experiments with $\alpha$ values set to $\{0.6, 0.75, 0.9, 0.95\}$. To gain insights into the underlying behavior, we analyze the transitions collected by the policy. The results can be found in Figure \ref{fig:rl-alpexp-alpha}.

From these results, we observe two behaviors depending on the values of $\alpha$. 
On one hand, when $\alpha$ is smaller than $0.9$, the RL policy demonstrates a periodic behavior. This phenomenon becomes more pronounced as $\alpha$ decreases, leading to instability that hinders WorldLLM's ability to improve consistently.

It seems that the reward diminishes to zero before the Scientist and Statistician can refine the existing hypothesis. This decline ultimately leads to a collapse in the RL agent's policy. As a result, the distribution of the collected data begins to resemble that of a random agent, as in the first iterations of WorldLLM. In response, the Scientist generates hypotheses that cause the framework to revert to an earlier stage. 

On the other hand, as $\alpha$ gets closer to 1, the reward takes increasingly longer to fade. This can be problematic as the reward function used is an estimation of the learning progress. For example, the high rewards collected at the beginning of training are no longer relevant after multiple iterations. 

This hyperparameter search explicitly shows that RL-ALPEXP has not fully resolved the sparse reward problem inherent to RL-ALP. If $\alpha$ is too small, the agent's policy collapses to random. If it is too close to $1$, the old rewards obtained require too much time to fade. 
One potential solution could be to add a small fraction of RL-LogP's rewards. This addition would prevent the agent's policy from collapsing but would require a new hyperparameter to balance the reward from RL-ALP and RL-LogP.

\end{document}